\documentclass[runningheads]{llncs}
\usepackage[T1]{fontenc}
\usepackage{graphicx}
\begin{document}
\title{Metacognitive AI: Framework and the Case for a Neurosymbolic Approach}
%
%
\author{Hua Wei\inst{1} \and
Paulo Shakarian\inst{1} \and
Christian Lebiere\inst{2} \and
Bruce Draper\inst{3} \and \\
Nikhil Krishnaswamy\inst{3} \and
Sergei Nirenburg\inst{4}}
\authorrunning{H. Wei et al.}
%
\institute{Arizona State University \email{\{hua.wei, pshak02\}@asu.edu}\\
\and
Carnegie Mellon University \email{cl@cmu.edu}\\
\and
Colorado State University \email{\{bruce.draper, nkrishna\}@colostate.edu, }\\
 \and
Rensselaer Polytechnic Institute \email{zavedomo@gmail.com}\\
}
\maketitle              
\begin{abstract}
Metacognition is the concept of reasoning about an agent’s own internal processes and was originally introduced in the field of developmental psychology.  In this position paper, we examine the concept of applying metacognition to artificial intelligence.  We introduce a framework for understanding metacognitive artificial intelligence (AI) that we call TRAP: transparency, reasoning, adaptation, and perception.  We discuss each of these aspects in-turn and explore how neurosymbolic AI (NSAI) can be leveraged to address challenges of metacognition.
\keywords{Metacognition  \and Neurosymbolic AI.}
\end{abstract}

\section{Introduction}

Metacognition is the concept of reasoning about an agent’s own internal processes and was originally introduced in the field of developmental psychology~\cite{flavell1979metacognition} as a description of higher-order cognition. This ``cognition about cognition'' is regarded by some as a self-monitoring process that is integral to the functioning of the human mind~\cite{demetriou1993architecture}. It has been studied extensively in the fields of manufacturing~\cite{li2017applications}, aerospace~\cite{izzo2019survey}, transportation~\cite{caesar2020nuscenes,abduljabbar2019applications,grigorescu2020survey}, and military applications~\cite{svenmarck2018possibilities}. We argue for the study of metacognitive artificial intelligence which deals with the reasoning about an artificial agent’s own processes. This idea actually has been studied on and off in the history of AI~\cite{cox2011metareasoning,cox2005metacognition}, but recent developments indicate that this area deserves a renewed focus. Specifically, despite large scale industry investments in AI, major failures still occur – which indicates that pure engineering solutions are unlikely to solve these fundamental failures. Consider the following examples:

\begin{itemize}
\item Large language model falsely accuses a professor of sexual harassment~\cite{Mok}. 
\item Autonomous robot taxi in San Francisco accidentally drags a woman for 20 feet causing major injury~\cite{Farivar_2023}.
\item Reinforcement learning model has to be retrained to play with slight changes in the environment~\cite{jayawardana2022impact,wei2022honor}.
\item Robot mistakes a man for a box in South Korea and crushes him to death~\cite{Atkinson_2023}. 
\end{itemize}

\begin{figure}
\centering
\includegraphics[width=0.5\columnwidth]{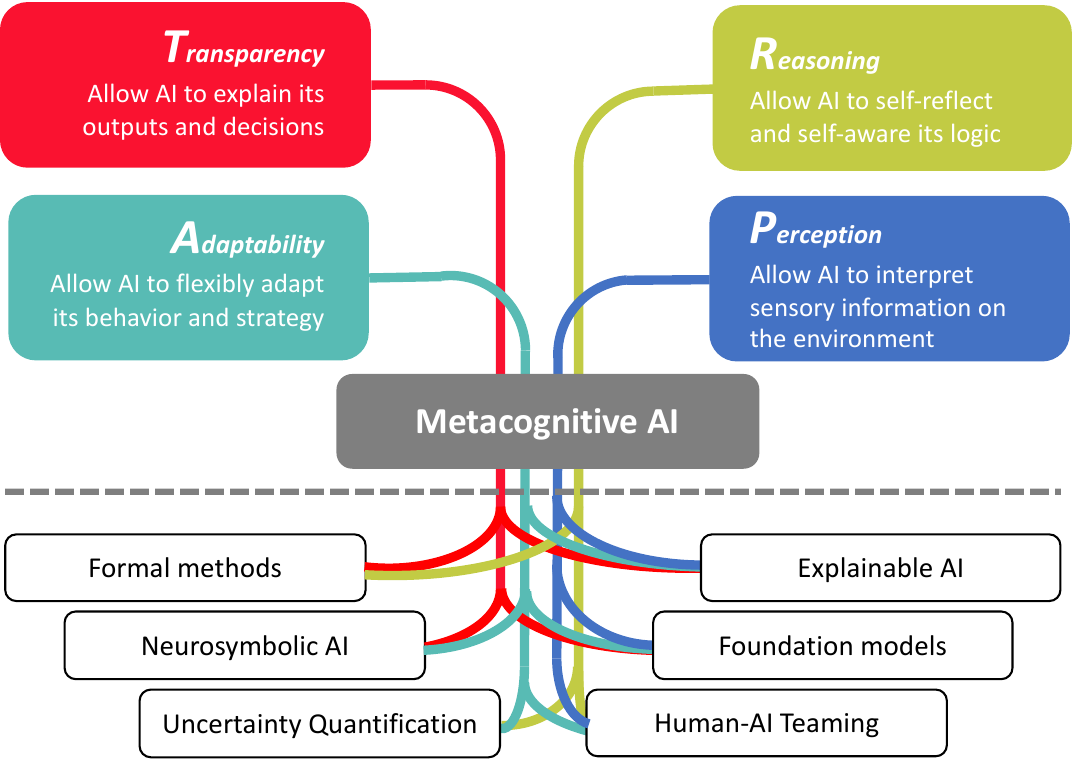}
\caption{Four aspects of metacognitive AI (TRAP) and approaches to achieve metacognition.}
\label{fig:intro:trap}
\end{figure}

Each of these case studies exhibits a different modality of AI failure. 
The first item illustrates a failure of \textbf{\underline{T}ransparency} – the system generated information that was false and could not provide a way to check itself on the facts. 
The second illustrates a failure in \textbf{\underline{R}easoning} – how the system synthesizes information and ultimately produces a decision. 
The third illustrates a failure of \textbf{\underline{A}daptation} – the system could not accommodate itself in a new environment.  
The fourth illustrates a failure in \textbf{\underline{P}erception} – how the system recognizes entities in its environment. 
In this introduction, which stemmed from the \emph{2023 ARO-sponsored Workshop on Metacognitive Prediction of AI Behavior}, we argue that the study of metacognitive AI should encompass these four areas (\textbf{\underline{TRAP}}), as is shown in Figure~\ref{fig:intro:trap}.  In this paper, we build on our recent ARO-sponsored workshop event~\cite{wei2024metaai} and examine each of these aspects and then discuss how neurosymbolic AI can be used as an approach to address challenges in metacognition.

\section{The TRAP Framework for Metacognition}
A traditional  artificial intelligence (AI) system could be simplified as $y=f_\theta(x)$, where $x$ is the input for the AI system; $y$, depending on applications, could be a description, prediction, or actions to take; and $f_\theta$ is the operational function in most AI systems with parameters $\theta$. A metacognitive AI system could be an additional function $g$. With different metacognitive areas, $g$ is in different locations concerning $f$.  With this framing in mind, we examine each aspect of the TRAP framework below.

\paragraph{Transparency} 
While traditional AI can sometimes be perceived as a `black box', metacognitive AI enhances trust and transparency by making the decision-making process in black-box AI more understandable to users. This is achieved through the function $g(f(x), \theta)$ or the function $g|f$. The function $g(f(x), \theta)$ represents the process of generating explanations based on both the input $x$ and the parameters $\theta$ of the model $f$, while the function $g|f$ represents the function $f$ with a series of $g$. This function allows metacognitive AI to explain its decisions in terms of both the input data and its internal parameters, catering to different user expectations and motivations for seeking explanations. 

On one hand, the nature of the explanation can vary significantly depending on whether it's intended for an expert with technical knowledge or a layperson. If they are looking to understand why certain outputs come from a global perspective, then the focus is to have $g(\theta)$ to make the $\theta$ transparent; if users are looking to understand certain cases, then the focus is to have $g(f(x))$ to understand a certain prediction $f(x)$ on $x$.  On the other hand, the purpose behind an explanation necessitates a different approach to how explanations are formulated and presented: is the explanation for enhancing the performance of the system, reducing bias, increasing fairness, or simply deriving a clearer understanding of the AI's decision-making process? Enhancing the performance of the system through transparency could involve using the explanations to correct predictions or induce actions $g|f$ where the understanding outputted from $g$ is then processed by the AI system to better learn $f$. 
~\cite{mitsopoulos2022toward} argued building cognitive models of both the AI and the human user that could be introspected upon to adapt explanations according to the discrepancy between the two models, e.g., when the AI decision does not conform to the human model's expectations.~\cite{wang2019designing} applied explainable AI framework to a real-world clinical machine learning (ML) use case, i.e., an explainable diagnostic tool for intensive care phenotyping. Co-designing with 14 clinicians, they provided five explanation strategies to mitigate decision biases and moderate trust. They implemented an early decision aid system to diagnose patients in an Intensive Care Unit (ICU) and found that users employed a diverse range of explainable AI facilities to reason.

\paragraph{Reasoning}
Traditional AI systems $f$ often rely on predefined algorithms and data sets for reasoning or decision-making, which can limit their effectiveness in dynamic or unfamiliar scenarios. In contrast, metacognitive AI incorporates self-reflection and self-awareness into its logic, represented by $f(x; g(\theta))$. This indicates how the AI's self-reflection (through $g$) informs its decision-making process (through $f$), enhancing its reasoning capabilities. For instance, a metacognitive AI in healthcare could use this approach to evaluate and refine its diagnostic criteria over time, learning to differentiate between complex cases and refining its diagnostic criteria based on outcomes. This leads to decisions that are not only based on data but also enriched by the AI’s growing experiential knowledge, resulting in more accurate and reliable outcomes.

In \cite{ulam2005using}, the authors showed that model-based reflection may guide reinforcement learning with two benefits: The first is a reduction in learning time as compared to an agent that learns the task via pure reinforcement learning. Secondly, the reflection-guided RL agent shows benefits over the pure model-based reflection agent, matching the performance of that agent in the metrics measured in addition to converging to a solution in fewer trials. In addition, the augmented agent eliminates the need for an explicit adaptation library such as is used in the pure-model-based agent and thus reduces the knowledge engineering burden on the designer significantly.
In \cite{andrychowicz2017hindsight}, a novel technique called Hindsight Experience Replay was introduced, whose intuition is to re-examine the trajectories with a different goal — while a trajectory may not help learn how to achieve the desired goal, it tells us something about how to achieve the state in the actual trajectory. They demonstrated this approach on the task of manipulating objects with a robotic arm on three different tasks: pushing, sliding, and pick-and-place, while the vanilla RL algorithm fails to solve these tasks.

\paragraph{Adaptability} 
Adaptation in metacognitive AI encapsulates the system’s ability to detect and correct errors of internal conditions and to flexibly adapt its behavior and strategies. This is represented by $f'(x; g(f(x), \theta))$, where $f'$ is the adapted model based on the metacognitive assessment $g$ of the original model's output $f(x)$ and parameters $\theta$. This notation reflects how metacognitive AI adapts by reassessing its outputs and parameters, allowing for more effective decision-making in the face of uncertainty and changing environments~\cite{da2023uncertainty,Ye_Chen_Wei_Zhan_2024}.
Additional adaptations could also be implemented with $g(x) ? f(x)\colon h(x)$, where the metacognitive process $g$ decides whether to use the main function $f$ or an alternative function $h$ based on its analysis of the input $x$. This could model AI systems that choose different processing paths based on metacognitive assessment without modifying $f$. Such systems can adapt to new environments and tasks by understanding their learning process and limitations. 
~\cite{leibig2017leveraging} showed that uncertainty-informed decision referral can improve diagnostic performance. 
More recently, another metacognitive approach allowing for adaptability known as \textit{error detection and correction rules} (EDCR) was introduced~\cite{xi2023rulebased}.  In this framework, function $g$ results in a set of learned rules that characterize the failure modes of $f$ and how to correct on those failure modes while $f'$ is an inference process conducted using these rules to erase or change the results of the underlying model $f$.  In \cite{xi2023rulebased} the authors applied this technique to the classification of geospatial movement trajectories and examined performance improvement on $f$, where the current state-of-the-art is neural architecture known as LRCN~\cite{kim2022gps}.  EDCR was able to both improve over the state-of-the-art as well as exhibit the ability to improve performance when exposed to out-of-distribution data.

\paragraph{Perception} 
Perception refers to the ability to interpret sensory information to understand the environment. Perception in metacognitive AI involves the system’s ability to interpret and understand sensory information, such as visual and auditory data, in a context-aware manner, represented by $f(g(x), x)$, where context is a metacognitive assessment of the AI's own capacity rather than an external context. Here, $f$ represents the primary perceptual processing function, interpreting sensory data like visuals or audio, while the metacognitive function $g$ specifically evaluates the accuracy and limitations of the AI's own sensory processing. The primary perceptual function $f$ then uses both the original sensory input $x$ and the metacognitive assessment $g(x)$ to refine its interpretation. This dual-input model allows the AI to recognize and compensate for any inherent biases or weaknesses in its perception.
This could include AI in autonomous vehicles that must interpret complex visual environments, in medical imaging distinguishing subtle diagnostic details, or in environmental monitoring systems that detect and analyze changes through sensory data.

\section{Neurosymbolic AI for Metacognition}

In the previous section, we introduced the TRAP framework of metacognitive AI , which provides a structured approach to understanding how metacognitive elements augment traditional AI systems. Building upon this foundation, in this section, we explore the emerging field of neurosymbolic AI (NSAI) and its profound implications for metacognition.   NSAI refers to the integration of connections (e.g., neural) with symbolic (e.g., logical) systems.  This term was coined in the early 2000's and has gained wider prominence in recent years~\cite{nsaiBook23,kautz22,thirdWave2020,DBLP:series/faia/369}.  The key relationships between NSAI and metacognition relate to the ability to use symbolic knowledge and perceptual models to detect and correct errors in each other (adaptability) and the use of symbolic languages to express information about error modes of a perceptual model (transparency).

With the introduction of Logic Tensor Networks~\cite{ltn22} the canonical paradigm for NSAI has consisted of guiding gradient descent with the addition of soft logical constraints in the loss function - and this was followed by related work~\cite{giunchiglia2020coheren,xu2018semantic}.  In general, these loss-based approaches would not fit the metacognitive paradigm, as in these incarnations of NSAI, the symbolic logic is used as an additional optimization criteria - in much the same way as one would a regularization term.  However, more recent views on NSAI do lend themselves to metacognition - in particular with respect to adaptability and transparency.

The key intuition in the use of NSAI for metacognitive adaptability is to leverage symbolic domain knowledge to explicitly identify errors in a neural model, allowing for some corrective action to be performed.  One well-known approach for NSAI metacognitive adaptability is abductive learning (ABL)~\cite{dai2019bridging}.  Using the paradigm of adaptability introduced in this paper, function $f'$ returns a result based on the combination of a perceptual model, a-priori domain knowledge (i.e., a logic program), and abduced error information (function $g$ in our framework).  Here, function $g$ is abduced based on inconsistencies between the perceptual model and domain knowledge and can take the form of additional symbolic structures added to the logic program and/or updates to the perceptual model ($f$ in our framework).  ABL has been shown to provide SOTA performance on combined perception-reasoning tasks as well as application to the identification of new concepts as shown in \cite{huang2023enabling}.  More recent applications of NSAI to metacognitive adaptability have sought to disentangle perceptual updates from the base perceptual model.  Specifically, \cite{cornelio2022learning} introduces a framework where an additional transformer model ($g$ in our framework) is used to predict errors in the underlying neural ($f$) using symbolic knowledge and reinforcement learning to detect and correct perceptual errors.  The work of \cite{xi2023rulebased} also addresses perceptual errors but using a rule-learning approach - here rules are learned about the results of the neural model that allow for error detection and correction while providing the byproduct of an explanation of the errors.

Complementary to the NSAI work relevant to metacognitive adaptability is NSAI work relating to metacognitive transparency.  Here, NSAI is used to reason directly about the inner workings of a perceptual model, often for a downstream task involving an explanation of the perceptual results.  One example of such work is \cite{EVANS2021103521} where a binarized neural network is used to produce a symbolic theory of perception used in a downstream task of appreciation, providing an explanation of the perceptual results.  Another application of NSAI to transparency deals with the use of concept induction~\cite{dalal2023explaining} to map activations in a neural network to an explanation using description logic - thereby providing transparency.

\section{Challenges}

The idea of metacognitive AI leads to many open questions.  In the application of NSAI to metacognition, there are several such as the challenge of creating symbolic structures to reason about (e.g., using inductive logic programming to obtain a knowledgebase~\cite{evans2018learning}, leveraging common sense knowledge~\cite{du-etal-2021-learning} - both still have major challenges).  More broadly, there are challenges that apply to metacognitive AI in general, which include the following:

\begin{itemize}
\item \textit{Generalization to Diverse Dynamic Environments.} Metacognitive AI must be capable of adapting to rapidly changing and unpredictable environments, or at least know when it is incapable.
\item \textit{Designing for Continuous Self-Improvement.} Enabling AI systems to not only identify their weaknesses or errors but also to autonomously modify their behavior and learning strategies for continuous improvement.
\item \textit{Ensuring Ethical and Responsible Metacognition.} As metacognitive AI systems will have a higher level of autonomy in decision-making, ensuring that these decisions are ethically and morally responsible.
\item \textit{Interpreting Metacognitive Processes.}  While explainability in AI is already a challenge, making the metacognitive processes of AI interpretable and understandable to humans adds an extra layer of complexity.
\item \textit{Benchmarking Datasets and Baselines.}  Such benchmarks would afford validation of an AI’s self-assessment and adaptive learning capabilities are functioning as intended, which requires human assessment. 
\end{itemize}
While the ideas of metacognitive AI are still very new, the ideas of error correction have previously been used to establish the foundations for technologies such as computer networking and digital signal processing.  We believe a formal study of the topic with respect to AI may yield similar advances in the future.

\begin{credits}
\subsubsection{\ackname} This work was funded by the Army Research Office (ARO) under grants W911NF2310345 and W911NF2410007.
\subsubsection{\discintname}
The authors have no competing interests.
\end{credits}

\bibliographystyle{splncs04}
\bibliography{bibtemp}

\end{document}